\documentclass{article}

     \usepackage[final]{neurips_2019}

\usepackage[utf8]{inputenc} 
\usepackage[T1]{fontenc}    
\usepackage{hyperref}       
\usepackage{url}            
\usepackage{booktabs}       
\usepackage{amsfonts}       
\usepackage{nicefrac}       
\usepackage{microtype}      
\usepackage{graphicx}
\usepackage{adjustbox}

\title{Fairness Assessment for Artificial Intelligence in Financial Industry}

\author{%
  Yukun Zhang\thanks{alternative email: yukunzhang329@gmail.com} \\
  ATB Financial\\
   Edmonton, AB T5J 0N3 \\
  \texttt{yzhang2@atb.com} \\
   \And
   Longsheng Zhou \\
   ATB Financial \\
   Edmonton, AB T5J 0N3 \\
  \texttt{lzhou2@atb.com} \\
}

\begin{document}

\maketitle

\begin{abstract}
 Artificial Intelligence (AI) is an important driving force for the development and transformation of the financial industry. However, with the fast-evolving AI technology and application, unintentional bias, insufficient model validation, immature contingency plan and other underestimated threats may expose the company to operational and reputational risks.  In this paper, we focus on fairness evaluation, one of the key components of AI Governance, through a quantitative lens. Statistical methods are reviewed for imbalanced data treatment and bias mitigation. These methods and fairness evaluation metrics are then applied to a credit card default payment example. 

\end{abstract}

\section{Introduction}


Financial intelligence has a fast and accurate machine learning capability to achieve the intellectualization, standardization, and automation of large-scale business transactions. Thus, it can improve service efficiency and reduce costs. For this reason, financial institutions are aggressively building their data scientist team to adopt AI/ML, however, the governance of AI/ML needs more attention.

AI presents particularly complex challenges to value-based decision-making because it introduces complexity, uncertainty, and scale. These features mean that harm is more challenging to evaluate and pre-empt, and that even the smallest harmful decision may have a large long-term effect that is difficult to detect and fix.

An AI/ML governance framework creates a robust business and operational framework to be utilized during the entire AI lifecycle and establishes common risk definitions and directions related to governance. The governance framework is essential for an organization to define safe boundaries for their data scientists. One of the biggest challenges in establishing governance framework is fairness assessment. The quantitative fairness assessment is at its early stage, and is challenging to be operationalized while technical standards and common technical best practices are still establishing among the data scientist teams. 

In the fairness assessment, data bias is sometimes the root cause of unintentional biases. At the same time, financial data is prone to bias and imbalance. The cost of having an unfair product or misclassifying a ''target'' event is high. These are ''features'' of financial data which need extra care when building models.

The challenges of biased and unbalanced data and the common methods of mitigating data bias and treating imbalanced data are reviewed in Section 2. Section 3 is an implementation of these methodologies to a credit card default payment dataset. 

\section{Challenges and Solutions on Modelling Financial Data}
\subsection{Challenges}
In financial institutions, because the data is collected from their customers, it is prone to bias and imbalance. For example, some world elite business credit cards have more male holders than female holders; the withdrawal and contribution to a registered retirement saving plan (RRSP) have strong age effect.  Here we separate the data challenges into data bias and imbalanced data, as they have different mitigation/treatment methods.

\subsubsection{Data Bias}

\paragraph{Gender Bias}
Historically, women and men have different roles in economies and societies. As employees, women are disproportionately concentrated in low paid, insecure, 'flexible' work, and in work that is - or is perceived to be-low-skilled (Staveren 2001).
Financial markets suffer from 'gender distortions' - distortions that disadvantage female borrowers as well as female savers, aside from the lack of collateral that limits women's access to finance (Baden 1996).
Fay and Williams (Fay 1993) presented that women can experience gender discrimination when seeking start-up capital. Similarly, Ongena and Popov (Ongena 2013) found firms owned by females have more difficulties obtaining credit than otherwise similar firms owned by males.  The female-owned firms do not underperform male-owned firms in terms of sales growth, even when not obtaining credit or when based in high gender-bias countries. With today's women earning higher incomes, playing a more active role in household financial decisions, and seeking additional professional investing advice, the issue of gender bias in financial services still exists.  Mullainathan et al. (Mullainathan 2012) found that, compared with male investors, female investors were less frequently asked about their personal and financial situation and more frequently advised to hold more liquidity, less international exposure, and fewer actively managed funds. Sahay et al. (Sahay 2018) found that greater inclusion of women as users, providers, and regulators of financial services would have benefits beyond addressing gender inequality. Narrowing the gender gap would foster greater stability in the banking system and enhance economic growth. It could also contribute to more effective monetary and fiscal policy.

\paragraph{Racial Bias}

Cohen-Cole (Cohen-Cole 2011) found qualitatively large differences in the amount of credit offered to similarly qualified applicants living in black versus white areas. Census (Federal Deposit Insurance Corporation 2014) data shows black and Hispanic Americans are more likely to go underbanked or deprived of conventional banking services than white or Asian Americans.
The census is not a biased result, however, the AI/ML can amplify this difference and then generate biased conclusions.
In the Fintech era, discrimination in lending can occur either in face-to-face decisions or in algorithmic scoring. Bartlett et al. (Bartlett 2019) found that lenders charge Latinx/African-American borrowers 7.9 and 3.6 basis points more for purchase and refinance mortgages respectively, costing them \$765M in aggregate per year in extra interest. FinTech algorithms also discriminate, but 40\% less than face-to-face lenders. 
\paragraph{Age Bias}
A 2002 study from the Federal Reserve Board (Braunstein 2002) showed that many people in underserved populations may be unfamiliar with components of the financial system. A combination of growing complexity, increases in consumer responsibility, as well as the noted changes in the structure of personal nuance to include more individual credit, have contributed to differences in financial literacy.
Cohen-Cole (Cohen-Cole 2011) also found age effects in his studies. For example, a senior may miss credit card payment because he or she forgets to do so, he or she uses the card irregularly, or doesn't understand the jargon on the credit card statement.

To summarize, it is important to mitigate gender, racial, and age biases when using AI solutions to solve business problems, which is beneficial to financial institutions and to eliminate bias in a broader context. Neglecting bias in the source data can cause bigger bias in the model conclusion, intentionally or unintentionally. In addition to mitigating bias, being bias-aware can help us build a customized solution. For example, for senior customers, we can identify their needs and provide special products.

\subsubsection{Imbalanced Data}
Imbalanced data set occurs when there is an unequal representation of classes. The data we collected can be imbalanced in two ways: one is due to the distribution of customers, the other one is due to the nature of the event. For the former one, gender bias is an example. Because we may have more male customers, female customer's financial habits are always predicted based on male customer's habits. The latter happens in certain areas such as fraud detection and risk management. The case of true fraudulent event is far less than legitimate cases, which cause data imbalance. The problem of imbalanced data is that since the probability of an event happening belongs to the majority class is significantly high, the algorithms are much more likely to classify new observations to the majority class. For example, if a lot of fraud events happen in Africa, then normal events happen in Africa will be marked as suspicious easier than events happen in North America. This way, the bias exists in the training set is amplified by the machine learning model. On the other hand, predictive models on imbalanced data is prone to high false negative rate, because the number of positive events is small. This is a big concern because the cost of false negative can be much higher than a false positive event. For example, in fraud detection, we would rather have a case identified as fraud but turns out to be legitimate, instead of missing a true fraudulent event.  

\subsection{Solutions}
\subsubsection{Bias Mitigation}
Bias mitigation algorithms can be categorized into pre-processing, in-processing (algorithm modifications), and post-processing. Table ~\ref{table1} shows the mitigation algorithms we capture in this paper. 
\begin{table}[]
  \caption{Bias mitigation algorithms}
  \label{table1}
\begin{adjustbox}{width=1\textwidth}
\begin{tabular}{|l|l|l|l|l|}
\hline
\textbf{Pre-processing}  & Reweighing                    & Optimized Preprocessing                  & Learning Fair Representations & Disparate Impact Remover      \\ \hline
\textbf{In-processing}   & Adversarial Debiasing          & Prejudice Remover                        &                               &                               \\ \hline
\textbf{Post-processing} & Equalized Odds Post-processing & Calibrated Equalized Odds Post-processing & Reject Option Classification  & Discrimination-Aware Ensemble \\ \hline
\end{tabular}
\end{adjustbox}
\end{table}

\textbf{\textit{Pre-processing algorithms}}
\paragraph{Reweighing}
Calders et al. (Calders 2009) proposed this method for the case where the input data contains unjustified dependencies between some data attributes and the class labels. The algorithm aims to reduce the dependence to zero while maintaining the overall positive class. Instead of relabeling the objects, different weights will be attached to them. According to these weights the objects will be sampled (with replacement) leading to a dataset without dependence (balanced dataset). On this balanced dataset the dependency-free classifier is learned.
\paragraph{Optimized preprocessing}
 (Calmon 2017) It probabilistically transforms the features and labels in the data with group fairness, individual distortion, and data fidelity constraints and objectives. This method also enables an explicit control of individual fairness and the possibility of multivariate, non-binary protected variables. 
 \paragraph{Learning fair representations}

 Zemel et al. (Zemel 2013) proposed this learning algorithm for fair classification that achieves both group fairness (the proportion of members in a protected group receiving positive classification is identical to the proportion in the population as a whole), and individual fairness (similar individuals should be treated similarly).  The algorithm encodes the data as well as possible, while simultaneously obfuscating any information about membership in the protected group.
 
  \paragraph{Disparate impact remover}(Feldman 2015) It edits feature values to mask bias while preserving rank-ordering within groups.
 
 \textbf{\textit{In-processing algorithms}}
\paragraph{Adversarial debiasing} (Zhang 2018) It is a framework for mitigating biases by including a variable for the group of interest and simultaneously learning a predictor and an adversary. It maximizes prediction accuracy and simultaneously reduce an adversary's ability to determine the protected attribute from the predictions. The approach is flexible and applicable to multiple definitions of fairness as well as a wide range of gradient-based learning models.
 \paragraph{Prejudice remover} (Kamishima 2012) This method adds a regularizer to the learning objective, which enforces a classifier's independence from sensitive information.
 
 \textbf{\textit{Post-processing algorithms}}

\paragraph{Equalized odds post-processing} (Hardt 2015) It solves a linear program to find probabilities with which to change output labels to optimize equalized odds. 
\paragraph{Calibrated equalized odds post-processing} (Pleiss 2017) This method optimizes over calibrated classifier score outputs to find probabilities with which to change output labels with an equalized odds objective. 
Kamiran et al. (Kamiran 2012) proposed the reject option of probabilistic classifier(s) (\textbf{Reject Option Based Classification}) and the disagreement region of general classifier ensembles (\textbf{Discrimination-Aware Ensemble}) to reduce discrimination. The former one gives the idea of a critical region in which instances belonging to deprived and favoured groups are labeled with desirable and undesirable labels, respectively. The latter makes an ensemble of (probabilistic, non-probabilistic, or mixed) classifiers discrimination-aware by exploiting the disagreement region among the classifiers.

\subsection{Fairness Metrics}

We select the following metrics to measure the model fairness: 
\paragraph{Statistical Parity Difference} This is the difference in the probability of favourable outcomes between the unprivileged and privileged groups. This can be computed both from the input dataset as well as from the dataset output from a classifier (predicted dataset). A value of 0 implies both groups have equal benefit, a value less than 0 implies higher benefit for the privileged group, and a value greater than 0 implies higher benefit for the unprivileged group. 
\paragraph{Equal Opportunity Difference} This is the difference in true positive rates between unprivileged and privileged groups. A value of 0 implies both groups have equal benefit, a value less than 0 implies higher benefit for the privileged group and a value greater than 0 implies higher benefit for the unprivileged group. 
\paragraph{Disparate Impact }This is the ratio in the probability of favourable outcomes between the unprivileged and privileged groups. This can be computed both from the input dataset as well as from the dataset output from a classifier (predicted dataset). A value of 1 implies both groups have equal benefit, a value less than 1 implies higher benefit for the privileged group, and a value greater than 1 implies higher benefit for the unprivileged group.

Specifically, we think equal opportunity is an important measurement of bias for financial institutions. For example, it is important to make sure customers know if they will pay back a loan, they will have the same chance of getting the loan with other applicants, regardless of their age, gender, and ethnicity.
\subsection{Bias Mitigation Tools}

Several open source libraries have become available in recent years, which make the goal of bias detection and mitigation easier to achieve. FairML (Adebayo 2016) is a toolbox for auditing predictive models by quantifying the relative significance of the inputs to a predictive model which can be used to assess the fairness (or discriminatory extent) of such a model. Fairness comparison (Friedler 2019) is an extensive library includes several bias detection metrics as well as bias mitigation methods, including disparate impact remover and prejudice remover mentioned above. AI Fairness 360 (AIF 360) (Bellamy 2018) is an open source toolkit that includes 71 bias detection metrics and 9 bias mitigation algorithms. Because of its comprehensiveness and usability, this paper uses AIF 360 for the case study.

\subsection{Imbalanced data treatment}
Feeding imbalanced data to a classifier can make it biased in favour of the majority class, simply because it did not have enough data to learn about the minority. 
One of the methods of treating imbalanced data is resampling, which includes under-sampling and over-sampling. Randomly removing instances from the majority class to achieve balance is called random under-sampling. Random over-sampling compensates the imbalanced class distribution by randomly replicating instances from the minority class. Under sampling can potentially lead to loss of information while oversampling can cause overfitting, as it makes exact replications of the minority samples rather than sampling from the distribution of minority samples. Synthetic Minority Over-sampling Technique (SMOTE) (Chawla 2002) is one of the most popular sampling methods for class imbalance. It is an over-sampling approach in which the minority class is over-sampled by creating ''synthetic'' examples rather than by over-sampling with replacement. These synthetic examples are created by a linear interpolation between a minority class instance and its nearest neighbours. 
The reasons of choosing SMOTE for our case study are 1) Choosing an over-sampling method avoid losing information. 2) Batista, Prati and Monard (Batista 2004) showed that SMOTE outperform several other (over and under) sampling methods 3) SMOTE overcomes the above mentioned issues.

\section{Case Study}
\subsection{Data}
A default credit card clients data set (Yeh 2009) from the UCI Machine Learning Repository (Bache 2013) is used in this case study. The outcome of this dataset is default payment (Yes=1, No=0). It includes 23 explanatory variables including amount of the given credit, gender (1=male, 2=female), education, marital status, age, history of past payment, amount of bill statement, and amount of previous payment. This data set collects information from 30,000 credit card clients in Taiwan from April 2005 to September 2005.
Yeh, the author of this data set, used it to compare the predictive accuracy of probability of default among six data mining methods.
\subsection{LightGBM Algorithm}
LightGBM (Ke 2017) is a gradient boosting framework that uses tree-based learning algorithms. It is used in this case study because of its widely recognized advantages of faster training speed and higher efficiency, lower memory usage, and better accuracy. In this case study, we use python library lightgbm by the same authors of algorithm LightGBM. 
\subsection{Model Fitting}
In the following different model fitting methods, the original data set is split into a training set and a test set in a ratio of 7:3; and a 5-fold cross-validation is used for model validation. A model using lightgbm is built to predict whether a customer will default or not. The focus of this case study is to see the impact of imbalanced data, usage of bias metrics, and how to remove biases. Parameter tuning and selection of modelling algorithms are out of our scope. For this purpose, this paper presents five models for this data set:  a lightgbm model with the original data, a lightgbm model with treated balanced data, a lightgbm model with bias removed data, a lightgbm model with treated balanced and bias removed data, and a set of manipulated data with bias mitigating algorithm applied. 

Confusion matrix together with precision, recall, f1-score are used to evaluate the performance of model fitting. Fairness indicators include statistical parity difference, equal opportunity difference, and disparate impact are used to evaluate the model fairness.
\subsubsection{Case 1: Plain LightGBM algorithm}
It is a common case that since most of the customers make payments to their credit cards, the number of default payment is much less. In this case, the overall no default payment vs. default payment in the original dataset is almost 4:1 (Table ~\ref{table2}). As mentioned above, the dataset is imbalanced. By fitting the data using lightGBM, we get the following result:
The overall accuracy of this model is 0.82 (Table ~\ref{table6}), but as expected for imbalanced data, the recall and F1-score is very low: 0.38 and 0.49, respectively. Because of the large size of the no default sample, the model tends to predict new cases towards no default, which results in a high false negative rate (0.62). 
\begin{table}[]
  \caption{Data distribution by age}
  \label{table2}
  \centering
\begin{tabular}{|l|l|l|l|l|l|}
\hline
        & \multicolumn{4}{c|}{Age}      &       \\ \hline
Default & 21-30 & 41-50 & 51-60 & 61-80 & Sum   \\ \hline
No      & 9530  & 4219  & 1276  & 164   & 15189 \\ \hline
Yes     & 2700  & 1302  & 437   & 52    & 4491  \\ \hline
\end{tabular}
\end{table}

\begin{table}[]
 \caption{Balanced data}
  \label{table3}
  \centering
\begin{tabular}{|l|l|l|l|l|l|l|}
\hline
        & \multicolumn{5}{c|}{Age}              &       \\ \hline
Default & 21-30 & 31-40 & 41-50 & 51-60 & 61-80 & Sum   \\ \hline
No      & 9530  & 8175  & 4219  & 1276  & 164   & 23364 \\ \hline
Yes     & 8862  & 9060  & 4357  & 987   & 98    & 23364 \\ \hline
\end{tabular}
\end{table}
\subsubsection{Case 2: Synthetic Balanced data}
To adjust the imbalanced data, here we use the above-mentioned SMOTE method to generate synthetic data for the default class. From Table ~\ref{table3}, we see the data is now balanced. Using lightgbm to fit the balanced data, we get an improved result.
The overall accuracy is 0.81, comparable to case 1, moreover, the performance metrics for the default group have been largely improved. The false negative rate is 0.23, much lower than the case 1 result. 
From the comparison to case 1, we see the importance of imbalanced data treatment. We now have the confidence of the predictions for both the majority and minority classes.

\subsubsection{Case 3: Bias Mitigated data}
After dealing with imbalanced data, we look into model fairness evaluation. The starting point of model fairness evaluation is evaluating data bias. Using the above-mentioned AIF 360 toolkit, we first look at the bias for unprivileged group (female, n=18112) and privileged group (male, n=11888). Note here that we are not sure whether female group is the unprivileged group or privileged one, just set it this way to begin the analysis. 

Here we choose to mitigate bias through pre-processing method reweighing. AIF 360 applied re-weighing by changing weights applied to training samples. Because the numbers of different gender groups are not strongly imbalanced, we don't use the data balancing technique towards this variable. 

\begin{table}[]
 \caption{Fairness metrics before and after mitigation}
  \label{table4}
  \centering
\begin{tabular}{|l|l|l|}
\hline
                         & Before & After \\ \hline
Diff. Statistical Parity & 0.0345 & 0.000 \\ \hline
Disparate impact         & 1.0457 & 1.000 \\ \hline
\end{tabular}
\end{table}
Table ~\ref{table4} shows before reweighing, male group was getting 3\% more positive outcomes (default payment) in the training dataset. This is not a huge difference between the two groups, but we still continue the analysis to show the effect of bias mitigation. After reweighing the training data, the differences between these two groups are eliminated.

\subsubsection{Case 4: Synthetic Balanced and Bias Mitigated data}

One step forward, we combine our experiments from case 2 and 3 and apply them together in case 4. By using SMOTE method to generate synthetic balanced data and use AIF 360 toolkit to mitigate bias in training data, we fit it with LightGBM again. 

\begin{table}[]
  \caption{Fairness metrics for all cases}
  \label{table5}
  \centering
\begin{tabular}{|l|c|c|c|l}
\hline
                             & \multicolumn{3}{c|}{Fairness}                                               \\ \hline
                             & Diff. statistical parity & Diff. equal opportunity & Diff. disparate impact \\ \hline
Case 1                    & 0.0213                   & 0.0113                  & 1.0246                 \\ \hline
Case 2                & 0.3034                   & 0.1439                  & 1.7955                 \\ \hline
Case 3            & 0.0168                   & 0.0099                  & 1.0194                 \\ \hline
Case 4 & 0.1621                   & -0.0104                 & 1.3688                 \\ \hline
\end{tabular}
\end{table}

\begin{table}[]
  \caption{Performance metrics for all cases}
  \label{table6}
  \centering
\begin{tabular}{|l|c|c|}
\hline
                             & \multicolumn{2}{c|}{Performance} \\ \hline
                             & Accuracy  & False Negative Rate  \\ \hline
Case 1                      & 0.82      & 0.62                 \\ \hline
Case 2                 & 0.81      & 0.23                 \\ \hline
Case 3          & 0.82      & 0.64                 \\ \hline
Case 4 & 0.83      & 0.19                 \\ \hline
\end{tabular}
\end{table}

Table ~\ref{table5} shows the model fairness measurements for all four models. We noticed that in the original data, the fairness differences between gender groups are not big, however, it is amplified by generating synthetic data (0.3034 versus 0.0213 for original data). So in case 4, we generate synthetic balanced data first and then run the bias mitigate process, aiming to reduce the bias. 
\begin{table}[]
  \caption{Performance metrics for case 4}
  \label{table7}
\centering
\begin{tabular}{|l|c|c|c|c|}
\hline
             &   Precision   &Recall      & F1-score     & Support      \\ \hline
No           & 0.79 & 0.84 & 0.82 & 6989  \\ \hline
Yes          & 0.83 & 0.78 & 0.81 & 7030  \\ \hline
Accuracy     &      &      & 0.82 & 14019 \\ \hline
Weighted Avg & 0.81 & 0.81 & 0.81 & 14019 \\ \hline
\end{tabular}
\end{table}
Table ~\ref{table5} shows that the fairness differences between gender groups are slightly reduced, yet still larger than the original dataset. According to difference in statistical parity, the probability of favourable outcome (no default) for female group is 0.1621 higher than the male group. In other words, male group is the unprivileged group--for a male customer, the probability of being predicted to be default is higher than a female customer.  

On the other hand, this model has the best performance in model accuracy, with a false negative rate of 0.19. From the model performance's perspective, the result of case 4 is similar to case 2. This shows the effect of balancing data is much higher than the effect of mitigating bias. One possible reason for this is the bias in the original data was big, thus the effect of mitigation is not significant, which won't change the result in testing data much. 

Case 4 results also show that the model performance is not compromised by the reweighing algorithm. 

\subsubsection{Case 5: Manipulated Biased Data}

Since the bias isn't significant in the original data, we have an idea to manipulate the data to create bias by adding 30\% samples which all of them are male customers who have payment default.

\begin{table}[]
  \caption{Fairness metrics for manipulated data}
  \label{table8}
\centering
\begin{tabular}{|l|c|c|c|}
\hline
                    & \multicolumn{1}{l|}{Diff. statistical parity} & \multicolumn{1}{l|}{Diff. equal opportunity} & \multicolumn{1}{l|}{Diff. disparate impact} \\ \hline
Manipulated data    & 0.5614                                        & 0.5477                                       & 2.687                                       \\ \hline
Bias mitigated data & 0.03178                                       & 0.0079                                       & 1.0415                                      \\ \hline
\end{tabular}
\end{table}

Table ~\ref{table8} shows the manipulated data is largely biased towards female, which is as expected. We then applied reweighing method and the bias is successfully mitigated. This example shows the effect of bias mitigation. 

\subsection{Limitation and Discussion}
Only gender fairness is measured in the above cases, fairness assessment in age groups can also be measured in future studies. The numbers in 51-60 and 61-80 age groups are much smaller than the other groups. It will be interesting to see how does this impact the model performance and fairness.

 The reweighing algorithm didn't significantly reduce the bias in the results in test set. In future studies, we can look at other preprocessing algorithms, in-processing algorithms, and post-processing algorithms. 
 
Other model fitting algorithms can be applied, especially artificial neural networks, which Yeh et.al. used in their research, reported a R-squared of 0.965. It's also worth to try other algorithms to see which ones are not so sensitive to imbalanced data/biased data.
The five models in the case study session show the importance of balancing the data set for the purpose of reducing false positive rate and thus get a better model. However, it may at the same time amplify the biases. Model creators can choose which model to use, according to the need. If it is a credit card default prediction, it may be important not to miss the people who will actually default. Meanwhile, it is good to be aware of the potential bias of the model. The whole point of the fairness assessment is discovering and measuring potential concerns that require further scrutiny. 

\section{Conclusion}
With the increasing attention on model governance and more fairness evaluation and mitigation tools becoming available, the AI/ML solutions built in the financial industry will gain more trust from the customers and thus become more beneficial to the industry. In this paper, we review the challenges and methodologies in imbalance data treatment and fairness evaluation. Fairness metrics of models are important results, especially for unintentional biases. The metrics are helpful for making next step decisions to further reduce biases. In future studies, we will present a more sophisticated algorithm to fit the data and reduce biases. Model explainability is another important topic we would like to discuss in future studies, especially about how to maintain the balance between model accuracy and explainability.


\section*{References}

\small

[1] Adebayo, Julius A. 2016. FairML: ToolBox for diagnosing bias in predictive modeling. PhD diss., Massachusetts Institute of Technology.

[2] Bache, Kevin, and Moshe Lichman. 2013. UCI Machine Learning Repository, University of California. [http://archive. ics. uci. edu/ml].

[3] Baden, Sally. 1996. "Gender issues in financial liberalisation and financial sector reform." 2006.

[4] Bartlett, Robert, Adair Morse, Richard Stanton, and Nancy Wallace. 2019. "Consumer-lending discrimination in the FinTech era." National Bureau of Economic Research w25943.

[5] Batista, Gustavo EAPA, Ronaldo C. Prati, and Maria Carolina Monard. 2004. "A study of the behavior of several methods for balancing machine learning training data." ACM SIGKDD explorations newsletter 6. no.1 20-29.

[6] Bellamy, Rachel KE, Kuntal Dey, Michael Hind, Samuel C. Hoffman, Stephanie Houde, Kalapriya Kannan, Pranay Lohia et al. 2018. "AI fairness 360: An extensible toolkit for detecting, understanding, and mitigating unwanted algorithmic bias." arXiv 1810.01943.

[7] Braunstein, Sandra, and Carolyn Welch. 2002. "Financial literacy: An overview of practice, research, and policy." Fed. Res. Bull. 88:445.

[8] Calders, Toon, Faisal Kamiran, and Mykola Pechenizkiy. 2009. "Building classifiers with independency constraints." 2009 IEEE International Conference on Data Mining Workshops 13-18.

[9] Calmon, Flavio, Dennis Wei, Bhanukiran Vinzamuri, Karthikeyan Natesan Ramamurthy, and Kush R. Varshney. 2017. "Optimized pre-processing for discrimination prevention." Advances in Neural Information Processing Systems 3992-4001.

[10] Chawla, Nitesh V., Kevin W. Bowyer, Lawrence O. Hall, and W. Philip Kegelmeyer. 2002. "SMOTE: synthetic minority over-sampling technique." Journal of artificial intelligence research 16:321-357.

[11] Cohen-Cole, E. 2011. "Credit card redlining." Review of Economics and Statistics 93(2), 700-713.

[12] Fay, Michael, and Lesley Williams. 1993. "Gender bias and the availability of business loans." Journal of Business Venturing 8, no. 4: 363-376.

[13] Federal Deposit Insurance Corporation. 2014. 2015: FDIC national survey of unbanked and underbanked households. Census, Federal Deposit Insurance Corporation.

[14] Feldman, Michael, Sorelle A. Friedler, John Moeller, Carlos Scheidegger, and Suresh Venkatasubramanian. 2015. "Certifying and removing disparate impact." 21th ACM SIGKDD International Conference on Knowledge Discovery and Data Mining. ACM. 259-268.

[15] Friedler, Sorelle A., Carlos Scheidegger, Suresh Venkatasubramanian, Sonam Choudhary, Evan P. Hamilton, and Derek Roth. 2019. "A comparative study of fairness-enhancing interventions in machine learning." Conference on Fairness, Accountability, and Transparency. ACM. 329-338.

[16] Kamiran, Faisal, Asim Karim, and Xiangliang Zhang. 2012. "Decision theory for discrimination-aware classification." IEEE 12th International Conference on Data Mining. IEEE. 924-929.

[17] Kamishima, Toshihiro, Shotaro Akaho, Hideki Asoh, and Jun Sakuma. 2012. "Fairness-aware classifier with prejudice remover regularizer." In Joint European Conference on Machine Learning and Knowledge Discovery in Databases. Berlin: Springer. 35-50.

[18] Ke, Guolin, Qi Meng, Thomas Finley, Taifeng Wang, Wei Chen, Weidong Ma, Qiwei Ye, and Tie-Yan Liu. 2017. "Lightgbm: A highly efficient gradient boosting decision tree." Neural Information Processing Systems. 3146-3154.

[19] Mullainathan, Sendhil, Markus Noeth, and Antoinette Schoar. 2012. "The market for financial advice: An audit study." National Bureau of Economic Research w17929.

[20] Ongena, Steven, and Alexander A. Popov. 2013. "Take Care of Home and Family, Honey, and Let Me Take Care of the Money-Gender Bias and Credit Market Barriers for Female Entrepreneurs." European Banking Center Discussion Paper 2013-001.

[21] Pleiss, Geoff, Manish Raghavan, Felix Wu, Jon Kleinberg, and Kilian Q. Weinberger. 2017. "On fairness and calibration." In Advances in Neural Information Processing Systems 5680-5689.

[22] Sahay, Ms Ratna, and Mr Martin Cihak. 2018. "Women in Finance: A Case for Closing Gaps." International Monetary Fund. 

[23] Staveren, Irene van. 2001. "Gender biases in finance." (Gender \& Development 9) no. 1 : 9-17.

[24] Yeh, I-Cheng, and Che-hui Lien. 2009. "The comparisons of data mining techniques for the predictive accuracy of probability of default of credit card clients." Expert Systems with Applications 362: 2473-2480.

[25] Zemel, Rich, Yu Wu, Kevin Swersky, Toni Pitassi, and Cynthia Dwork. 2013. "Learning fair representations." International Conference on Machine Learning 325-333.

[26] Zhang, Brian Hu, Blake Lemoine, and Margaret Mitchell. 2018. "Mitigating unwanted biases with adversarial learning." 2018 AAAI/ACM Conference on AI, Ethics, and Society. ACM. 335-340.

\end{document}